\documentclass[runningheads]{llncs}
\usepackage[T1]{fontenc}
\usepackage{graphicx}
\usepackage{epstopdf}
\usepackage{multirow}
\usepackage{algorithm}
\usepackage{algorithmic}
\usepackage{amsmath}
\usepackage{amssymb}
\usepackage{subcaption}
\usepackage{dsfont}
\usepackage{mathrsfs}
\usepackage[dvipsnames]{xcolor}
\usepackage{tikz}
\usepackage{url}
\usepackage{xspace}
\DeclareMathOperator*{\argmin}{argmin}

\newcommand{\THEMES}{THEMES\xspace}

\begin{document}
\title{Hierarchical Apprenticeship Learning from  Imperfect Demonstrations with Evolving Rewards}




%
\titlerunning{HALIDE: Hierarchical Apprenticeship Learning Framework}
%
\author{Md Mirajul Islam\orcidID{0000-0001-5086-7399} \and
Rajesh Debnath\orcidID{0000-0002-5460-9773} \and
Adittya Soukarjya Saha\orcidID{0000-0001-6344-9663} \and
Min Chi\orcidID{0000-0003-1765-7837}}
\authorrunning{M. M. Islam et al.}
%
\institute{North Carolina State University, Raleigh, NC 27606, USA\\
\email{\{mislam22, rdebnat, asaha4, mchi\}@ncsu.edu}}
%
\maketitle              
\begin{abstract}

While apprenticeship learning has shown promise for inducing effective pedagogical policies directly from student interactions in e-learning environments, most existing approaches rely on optimal or near-optimal expert demonstrations under a fixed reward. Real-world student interactions, however, are often inherently imperfect and evolving: students explore, make errors, revise strategies, and refine their goals as understanding develops.  In this work, we argue that imperfect student demonstrations are not noise to be discarded, but structured signals—provided their relative quality is ranked. We introduce \textbf{HALIDE, Hierarchical Apprenticeship Learning from Imperfect Demonstrations with Evolving Rewards}, which not only leverages suboptimal student demonstrations, but \emph{ranks} them within a hierarchical learning framework. HALIDE models student behavior at multiple levels of abstraction, enabling inference of higher-level intent and strategy from suboptimal actions while explicitly capturing the temporal evolution of student reward functions. By integrating demonstration quality into hierarchical reward inference, HALIDE distinguishes transient errors from suboptimal strategies and meaningful progress toward higher-level learning goals. Our results show that HALIDE more accurately predicts student pedagogical decisions than approaches that rely on optimal trajectories, fixed rewards, or unranked imperfect demonstrations. 

\end{abstract}
%
%
\section{Introduction}
Most STEM e-learning environments, including Intelligent Tutoring Systems (ITSs), rely on \emph{pedagogical policies} to guide instructional actions, determining not only what content is presented but also when and how interventions are delivered throughout a student’s learning trajectory. These policies operate in a sequential decision-making setting, where each pedagogical choice influences future learning opportunities and long-term outcomes. Prior work has shown that Reinforcement Learning (RL) and Deep RL can induce data-driven pedagogical policies that outperform manually designed heuristics in controlled educational settings \cite{mandel2014offline,rowe2015improving,shen2016reinforcement}. However, these real-world RL and DRL adoption is limited by the large amount of data required to explore intricate learning environments and the inherent difficulty of manually engineering reward functions that accurately reflect sophisticated, long-horizon educational goals.\cite{kaelbling1996reinforcement,amodei2016concrete}.

\textbf{Apprenticeship Learning (AL)} provides an alternative by deriving policies from a limited set of \emph{expert demonstrations} through inference of the underlying reward function \cite{abbeel2004apprenticeship,ziebart2008maximum}. By assuming experts act optimally or near-optimally, AL can efficiently approximate effective decision-making. Student interactions in e-learning environments, however, are rarely optimal. Students actively explore different strategies, learn from mistakes, and refine their approaches, adapting their learning goals as they gain experience. This process reflects the presence of evolving internal reward functions that shift as students gain expertise and familiarity with the system. Conventional AL frameworks treat suboptimal actions as noise, failing to capture the dynamics of learning. We argue that imperfect student demonstrations instead constitute structured signals that can reflect evolving competencies, strategies, and reward priorities. Crucially, leveraging these signals requires more than simply incorporating suboptimal trajectories: their \emph{relative quality must be explicitly ranked}, as treating all imperfect demonstrations as equally informative obscures meaningful pedagogical structure.  

We introduce \textbf{HALIDE, Hierarchical Apprenticeship Learning from Imperfect Demonstrations with Evolving Rewards}, a framework that hierarchically models student behavior, explicitly captures the temporal evolution of goals while ranking demonstrations according to pedagogical quality. HALIDE infers high-level intent and strategy from suboptimal actions, separates transient errors from meaningful progress, and explicitly captures the temporal evolution of student goals. This perspective aligns with advances in imitation learning from noisy or mixed-quality demonstrations using confidence scores, rankings, or preference-based supervision \cite{zolna2020oril,wu2019imperfectil,tangkaratt2021robust,xu2021discriminatorweighted,liu2025pngail,hoang2024sprinql}. However, most methods operate in flat policy spaces, neglecting hierarchical decision-making and temporally evolving strategies critical to educational domains. Our prior work introduced \THEMES \cite{yang2025themes}, a hierarchical AL framework that models evolving reward functions via sub-trajectory partitioning and induce cluster-specific policies. While effective, \THEMES assumes that all input demonstrations are expert-level and equally informative, making it sensitive to noise arising from heterogeneous decision quality or suboptimal trajectories. HALIDE extends \THEMES with \emph{ranked hierarchical apprenticeship learning}, integrating continuous demonstration quality signals into hierarchical policy induction. Higher-quality pedagogical decisions exert greater influence, while the full trajectory structure is preserved. Crucially, ranking is embedded within the hierarchical learning loop, allowing high-level strategies to emerge from reward-regulated refinement rather than collapsing learning into a flat weighted imitation objective. 
 
 It is important to note that ranking in AL serves a fundamentally different role from reward functions in  RL and deep RL. In RL, rewards define the objective of behavior: they specify what the agent should optimize through interaction with the environment. Reward functions shape long-term return and drive policy improvement through trial-and-error exploration. Designing such rewards in educational settings is notoriously difficult, as pedagogical objectives are multi-faceted, long-horizon, and often only indirectly observable.  In contrast, ranking in AL does not define what behavior is optimal; instead, it modulates how much influence each demonstrated behavior should have when inducing a policy from observational data. Ranking answers a different question than rewards: rather than asking \textit{“What should be optimized?”}, it asks \textit{“Which observed decisions should be trusted more during learning?”} This distinction is critical in demonstration-driven settings.

We evaluate HALIDE against seven competitive AL baselines that rely on expert trajectories or fixed rewards, along with two ablation variants, on a challenging cross-semester prediction task. Models are trained on student trajectories from past semesters and tested on a subsequent semester to predict pedagogical decisions in an ITS. Our results showed that HALIDE consistently outperforms all baselines and ablations across evaluation metrics. 
\textbf{This work makes the following key contributions} :
1)  We introduce \textbf{HALIDE}, a hierarchical AL framework that models student pedagogical behavior from \emph{imperfect demonstrations} while explicitly capturing \emph{evolving reward structures}, enabling the inference of higher-level intent and strategy as student goals and competencies change over time.
2) We show that the key is to introduce quality-aware hierarchical apprenticeship learning that ranks demonstrations rather than treating them uniformly. 
3) Extensive cross-semester evaluation on a real-world ITS shows that HALIDE consistently outperforms flat and hierarchical AL baselines, with strong gains from suboptimal student demonstrations.

\section{Related Work}
\vspace{-0.1in}
\subsubsection{Pedagogical Policy Induction in ITSs}
A large body of work has explored the use of reinforcement learning (RL) and deep reinforcement learning (DRL) to induce pedagogical policies in intelligent tutoring systems \cite{mandel2014offline,rowe2015improving,hostetter2025human}. These approaches typically model tutoring as a sequential decision-making problem in which the tutor selects instructional actions to maximize long-term learning outcomes \cite{shen2016reinforcement,zhou2019hierarchical,tithi2026adaptive}.
While RL-based methods have demonstrated strong empirical performance, they often require large amounts of interaction data and rely on carefully designed reward functions, such as normalized learning gain, to guide optimization. In practice, such data is difficult to obtain without deploying exploratory or randomized tutoring strategies, limiting scalability and adoption in real classrooms. Apprenticeship learning (AL) offers an alternative by inferring pedagogical policies directly from demonstrated behavior rather than explicit reward specification \cite{abbeel2004apprenticeship,ziebart2008maximum}.
In ITS settings, AL is particularly appealing because it enables fully offline learning from historical interaction logs. However, unlike tutor-driven RL, apprenticeship learning in ITSs treats students as the source of pedagogical decisions, using their choices—such as when to request hints or view worked examples—as demonstrations \cite{mirajul2024emedm,islam2025generalized}. This shift introduces student agency into policy induction, but also raises new challenges related to variability in demonstration quality.

\vspace{-0.2in}
\subsubsection{Apprenticeship Learning from Mixed-Quality Demonstrations}
\label{subsec:rw_mixed_quality}

Imitation and apprenticeship learning traditionally assume access to high-quality expert demonstrations, but this assumption is often violated in real-world domains where expert data are scarce and heterogeneous trajectories are abundant.
While early offline approaches such as behavioral cloning are simple and effective with sufficient expert coverage, they are known to suffer from compounding error and distribution shift under limited or imperfect demonstrations \cite{SCHAAL1999233,pmlr-v15-ross11a}. To address this challenge, a growing body of work studies learning from \emph{mixed-} or \emph{imperfect-quality} demonstrations. Prior methods incorporate confidence or reliability signals to reweight demonstrations or state--action pairs, including confidence-weighted imitation learning \cite{wu2019imperfectil}, discriminator-based reweighting \cite{xu2021discriminatorweighted}, positive-unlabeled formulations \cite{liu2025pngail}, and value- or reward-based approaches that integrate expert and suboptimal experience \cite{zolna2020oril,hoang2024sprinql}.
These techniques improve robustness by attenuating the influence of unreliable behavior without requiring online interaction.

However, most existing approaches operate in \emph{flat} policy spaces and apply quality at a coarse granularity, such as trajectory-level weighting.
They do not account for hierarchical or temporally evolving decision-making, where the reliability of demonstrated actions may vary over time.
Our work complements this literature by incorporating continuous quality signals into a \emph{hierarchical} apprenticeship learning framework, enabling quality-aware policy induction while preserving temporal abstraction and evolving pedagogical strategies.

\vspace{-0.2in}
\subsubsection{Hierarchical Apprenticeship Learning and Evolving Strategies: }
\label{subsec:rw_hierarchical}
Hierarchical and multi-intent apprenticeship learning has been studied in settings where demonstrations are governed by multiple latent reward functions or intentions, often via Bayesian or EM-based formulations that cluster trajectories and learn cluster-wise rewards or policies \cite{babes2011apprenticeship,dimitrakakis2011bayesian,choi2012nonparametric}. 
Separately, hierarchical and multi-modal imitation learning approaches aim to capture sub-task structure and temporally extended behavior \cite{hausman2017multi,krishnan2016hirl}. In educational domains, modeling evolving strategies is particularly important because students' goals and learning needs change over time. \THEMES provides a time-aware hierarchical apprenticeship learning approach that models evolving reward functions by partitioning trajectories into sub-trajectories and inducing cluster-specific policies \cite{yang2025themes}. 
While effective at capturing temporal abstraction and strategic evolution, the framework assumes that demonstrations admitted into training are uniformly reliable, limiting robustness when demonstrations are mixed-quality or when reliability varies within trajectories. Our work extends this line of research by incorporating continuous quality signals into hierarchical policy induction, enabling hierarchical apprenticeship learning to better handle realistic mixed-quality student demonstrations.

\begin{figure}
    \centering
    \includegraphics[scale=0.4]{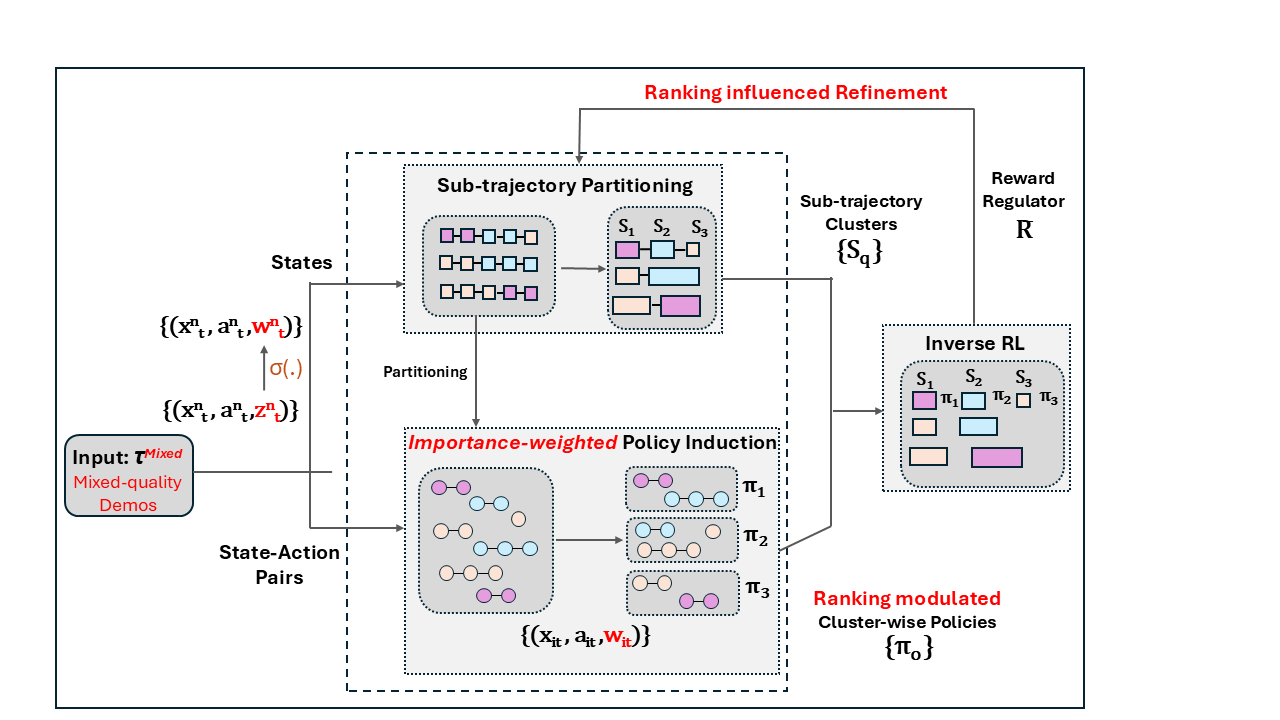}
    \caption{Overview of the HALIDE framework.}
    \label{fig:overview_ranked}
    \vspace{-0.12in}
\end{figure}

\section{Methods}
\label{sec:method}
The overview of HALIDE is illustrated in Figure~\ref{fig:overview_ranked}, with details provided in Algorithm~\ref{alg:halide_impl}. The input consists of $N$ demonstrated trajectories of mixed-quality,
$\tau^n=\{(\mathbf{x}_t^n, a_t^n)\}_{t=1}^{T^n}$, where $\mathbf{x}_t^n \in \mathbb{R}^m$ denotes the state at time $t$, $a_t^n \in \mathcal{A}$ denotes the pedagogical action chosen at time $t$, $T^n$ is the trajectory length, and $n \in [1, N]$. HALIDE augments each demonstrated decision with a continuous importance weight $w_t^n \in (0,1]$, yielding weighted demonstrations $\{(\mathbf{x}_t^n, a_t^n, w_t^n)\}_{t=1}^{T^n}$. Importantly, $w_t^n$ modulates the influence of demonstrated decisions during policy induction and does not define a reward function or alter the underlying trajectories.

\noindent \textbf{Notation.}
We use $Q$ to denote the number of high-level state clusters produced by RMT--TICC and $O$ to denote the number of high-level policy (action) clusters induced by EM--EDM. The resulting set of sub-trajectories after partitioning is denoted as $\{\xi_i\}_{i=1}^{\hat{N}}$, where $\xi_i=\{(\mathbf{x}_{it}, a_{it}, w_{it})\}_{t=1}^{T_i}$.

\begin{algorithm}[t]
\caption{\textbf{HALIDE}}
\label{alg:halide_impl}
\renewcommand{\thealgorithm}{}
\begin{algorithmic}[1]
    \STATE \textbf{Input:} trajectories $\{\tau^n\}_{n=1}^{N}$, quality signals $\{z_t^n\}$
    \STATE \textbf{Hyperparameters:} $Q,O,\omega,K,\alpha$
    \STATE Initialize $\overline{R}^{(0)}(\cdot)\gets 1$, iteration $k\gets 0$
    \WHILE{$k < K$}
        \STATE Run RMT--TICC using $\overline{R}^{(k)}$ to obtain $\{S_q^{(k)}\}$ and sub-trajectories $\{\xi_i^{(k)}\}$
        \STATE Run weighted EM--EDM on $\{\xi_i^{(k)}\}$ to obtain $\{\pi_o^{(k)}\}$ and $\{u_{io}^{(k)}\}$
        \STATE Learn $\overline{R}^{(k+1)}(\cdot)$ from high-level state-action pairs
        \STATE $k\gets k+1$
    \ENDWHILE
    \STATE \textbf{Output:} $\{\pi_o\}_{o=1}^{O}$
\end{algorithmic}
\end{algorithm}

\vspace{-0.2in}
\subsection{Demonstration Quality and Weight Construction}
\label{subsec:weight_construction}

HALIDE assumes access to continuous demonstration quality $z_t^n$ aligned with demonstrated decisions. This quality measure may originate from learning outcomes, confidence estimates, or other domain-specific evaluations. We map the continuous quality to a relative ranking signal or importance weight via:
\begin{equation}
\label{eq:weight_map}
w_t^n = \sigma(\alpha z_t^n) = \frac{1}{1+\exp(-\alpha z_t^n)},
\end{equation}
where $\alpha>0$ controls the sensitivity. When $z_t^n$ is computed at the trajectory level, it is distributed across time to obtain a coarse per-step proxy. These weights are used only to modulate policy induction; they do not alter the underlying trajectories or the segmentation procedure.

\vspace{-0.2in}
\subsection{High-level States Extraction (RMT--TICC)}
\label{subsec:rmt-ticc}
HALIDE adopts the same high-level state extraction module as \THEMES. Each state $\mathbf{x}_t^n \in \mathbb{R}^m$ is assigned to one of $Q$ clusters by considering a sliding window of size $\omega \ll T^n$. For each time step $t$, the context window is $\mathbf{X}_t^n=\{\mathbf{x}_{t-\omega+1}^n,\ldots,\mathbf{x}_t^n\}$, treated as an $m\omega$-dimensional random variable and clustered into $Q$ Gaussian components.

Reward-regulated Multi-series Time-aware Toeplitz
Inverse Covariance-based Clustering (RMT-TICC) incorporates time-awareness and decision-making patterns via a reward-regulated consistency constraint:
\begin{align}
\small
\label{eq:rmt_ticc_obj}
\argmin_{\mathbf{\Theta}, \mathbf{S}} \sum_{q=1}^{Q}
\bigg[
\sum_{n=1}^{N}\sum_{\mathbf{X}_t^n \in \mathbf{S}_q}
\Big( -\ell\ell(\mathbf{X}_t^n,\mathbf{\Theta}_q)
+ c(\mathbf{X}_{t-1}^n,\mathbf{S}_q,\Delta T_t^n,\overline{r}_t^n)\Big)
+ \lambda \|\mathbf{\Theta}_q\|_1
\bigg],
\end{align}
where $-\ell\ell(\cdot)$ is the negative log-likelihood term, $c(\cdot)$ is the reward-regulated time-aware consistency term, and $\|\mathbf{\Theta}_q\|_1$ imposes sparsity. $\Delta T^{n}_{t}$ denotes the interval between consecutive states, \textit{i.e.}, $\mathbf{X}_{t}^{n}$ and $\mathbf{X}_{t-1}^{n}$.

A high-level reward regulator $\overline{R}(\cdot)$ is learned from high-level state-action pairs and distributed to obtain $\overline{r}_t^n$, which regulates the consistency constraint. This hierarchical structure enables modeling evolving pedagogical strategies without requiring online interaction.
\subsection{High-level Actions Induction (Weighted EM--EDM)}
\label{subsec:weighted_emedm}

\noindent\textbf{EDM (strictly offline).}
Energy-based Distribution Matching (EDM) induces a stochastic policy $\pi^\theta$ from demonstrations by matching the demonstration occupancy measure $\rho_D$ with the induced occupancy $\rho_{\pi^\theta}$:
\begin{equation}
\small
\label{eq:edm_kl_halide}
\argmin_{\theta} D_{\mathrm{KL}}(\rho_D \,\|\, \rho_{\pi^\theta})
=
\argmin_{\theta} -\mathbb{E}_{(\mathbf{x},a)\sim \rho_D}\big[\log \rho_\theta(\mathbf{x},a)\big].
\end{equation}
This objective decomposes into (i) a behavior cloning term on demonstrated actions and
(ii) an occupancy regularization term involving the energy-model normalizer $U^\theta(\mathbf{x})$,
allowing policy learning in a strictly offline manner without environment interaction.

\noindent\textbf{EM--EDM over sub-trajectories.}
To model heterogeneous strategies and evolving rewards, EM--EDM assumes each sub-trajectory
$\xi_i=\{(\mathbf{x}_{it},a_{it})\}_{t=1}^{T_i}$ is generated by one of $O$ latent policy clusters,
with prior $\rho_o$ and parameters $\theta_o$.
Given current parameters, the E-step computes responsibilities using the standard sub-trajectory likelihood:
\begin{equation}
\small
\label{eq:emedm_likelihood_halide}
\Pr(\xi_i \mid \theta_o)\;\propto\;
\prod_{t=1}^{T_i}
\frac{\pi^{\theta_o}(a_{it}\mid \mathbf{x}_{it})}{U^{\theta_o}(\mathbf{x}_{it})},
\qquad
u_{io}
=
\frac{\rho_o \Pr(\xi_i\mid \theta_o)}
{\sum_{o'=1}^{O} \rho_{o'} \Pr(\xi_i\mid \theta_{o'})}.
\end{equation}
Importantly, this assignment step depends only on behavioral consistency and temporal structure,
and is not influenced by demonstration quality.

\noindent\textbf{HALIDE: decision-weighted objective in the M-step.}
HALIDE incorporates demonstration quality during policy induction by augmenting each sub-trajectory
with decision-level importance weights,
$\xi_i=\{(\mathbf{x}_{it}, a_{it}, w_{it})\}_{t=1}^{T_i}$,
where $w_{it}\in(0,1]$ reflects the relative reliability of the demonstrated decision at time $t$. In the M-step, HALIDE updates each policy cluster by minimizing a responsibility-weighted, decision-weighted objective:
\begin{equation}
\small
\label{eq:weighted_mstep_halide}
\argmin_{\{\theta_o\}}
\sum_{i=1}^{\hat{N}}\sum_{o=1}^{O} u_{io}
\sum_{t=1}^{T_i} w_{it}\,
\Big(-\log \pi^{\theta_o}(a_{it}\mid \mathbf{x}_{it})\Big)
\;+\;
\lambda
\sum_{i=1}^{\hat{N}}\sum_{o=1}^{O} u_{io}
\sum_{t=1}^{T_i} w_{it}\,
\log U^{\theta_o}(\mathbf{x}_{it}).
\end{equation}

Here, $w_{it}$ uniformly scales both the behavior cloning term and the EDM occupancy regularization,
ensuring that lower-quality decisions exert proportionally less influence during policy learning,
while preserving the strictly offline nature of EDM.
Responsibilities $u_{io}$ are computed from the unweighted likelihood in~\eqref{eq:emedm_likelihood_halide}.
When $w_{it}\equiv 1$ for all $(i,t)$, HALIDE reduces exactly to standard EM--EDM as used in \THEMES.

\vspace{-0.1in}
\subsection{High-level Reward Regulator and Effect of Ranking}
\label{sec:reward_regulator}
\vspace{-0.1in}
HALIDE maintains a high-level reward regulator $\overline{R}(\cdot)$ that captures evolving pedagogical intent across sub-trajectories.
The regulator is learned from high-level state--action statistics induced by the current hierarchical policies and modulates the temporal consistency term in RMT--TICC, biasing segmentation toward coherent pedagogical progress via EM-Inverse RL (EM-IRL). The objective function of EM-IRL is consistent with that of \THEMES \cite{yang2025themes}, with the primary difference being the inclusion of ranking signals. Ranking is applied only during low-level policy induction.
However, since the induced policies determine the state--action distributions used to estimate $\overline{R}(\cdot)$, ranking influences the hierarchy indirectly through an iterative feedback loop:
\[
\{w_{it}\}
\;\Rightarrow\;
\{\pi_o\}
\;\Rightarrow\;
\overline{R}(\cdot)
\;\Rightarrow\;
\text{refined sub-trajectory segmentation}.
\]
This coupling allows higher-quality decisions to shape high-level strategy discovery while preserving the hierarchical structure of learning.




\vspace{-0.2in}
\section{Experiments}
\begin{figure}[htbp]
    \centering
    \begin{minipage}[b]{0.55\textwidth}
        \centering
        \includegraphics[scale=0.2]{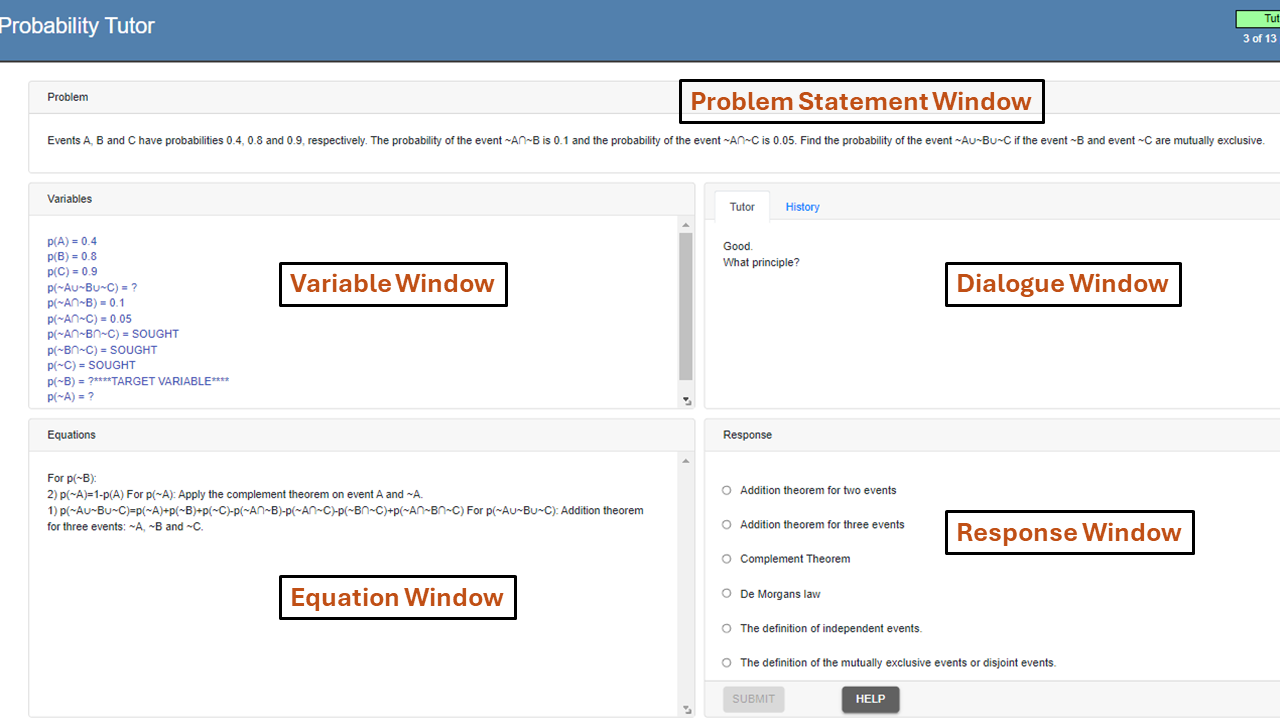}
        \caption{ITS Interface}
        \label{fig:pyrenees-interface}
    \end{minipage}
    \hfill
    \begin{minipage}[b]{0.42\textwidth}
        \centering
        \includegraphics[scale=0.2]{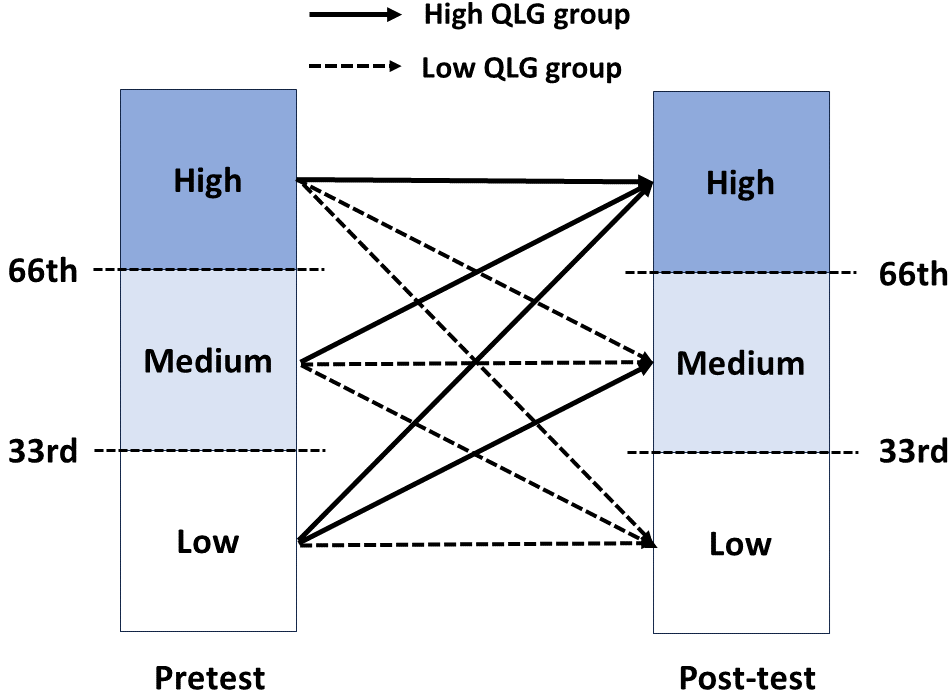}
        \caption{Quantized Learning Gain}
        \label{fig:qlg}
    \end{minipage}
    \vspace{-0.2in}
\end{figure}

\vspace{-0.1in}
This study utilized a Probability ITS (Figure \ref{fig:pyrenees-interface}) deployed across multiple Spring and Fall semesters of a large undergraduate STEM course at a public university. The system, developed by domain experts and overseen by departmental committees, is designed to support entry-level undergraduates in learning ten core probability principles, \textit{e.g.}, the complement theorem and Bayes’ rule, through a sequence of complex problem-solving activities.

\vspace{-0.1in}
\subsubsection{Data Collection}
Data were collected from 244 undergraduate students across five semesters: 57 students in Spring 2021 (S21), 56 in Spring 2022 (S22), 54 in Spring 2024 (S24), 44 in Fall 2024 (F24), and 33 in Spring 2025 (S25). All students followed an identical protocol consisting of a textbook study phase, a pre-test, ITS-guided training, and a post-test. The pre-test and post-test comprised 8 and 12 open-ended problem-solving questions, respectively, and were graded independently by two evaluators using a double-blind process, with discrepancies resolved through discussion. Scores were normalized to the range $[0,100]$. During ITS training, students interacted with the same 12 problems in a fixed order. For 10 of these problems, students exercised \emph{agency} by choosing among three pedagogical approaches: \emph{1) solving the problem independently}, \emph{2) collaborating with the ITS}, or \emph{3) viewing a worked-out example}. All interaction data were anonymized and obtained under an exempt IRB-approved protocol.

\vspace{-0.1in}
\subsubsection{State and Action Representation}
Students’ latent learning states are not directly observable. Following prior work, each interaction state is represented using a \emph{130-dimensional continuous} feature vector constructed by domain experts. These features capture multiple aspects of student behavior, including \textbf{\textit{10} Autonomy} features that measure the student's work, such as the number of elicits since the last tell; \textbf{\textit{22} Temporal} features that represent time-related information, like average time per step; \textbf{\textit{31} Problem Solving} features that reflect contextual information about problem-solving, including difficulty level; \textbf{\textit{57} Performance} features that indicate student's performance so far, such as the percentage of correct responses; \textbf{\textit{10} Hints} features that show data on hint usage, such as total hints requested. The action space consists of \emph{three pedagogical decisions} corresponding to the student’s choice of instructional support.

\vspace{-0.1in}
\subsubsection{Expert Demonstration Selection}
To identify optimal or near-optimal student demonstrations, we compute Quantized Learning Gain (QLG) measure \cite{mao2018tracing}, categorizing students into low, medium, or high-performance groups based on their pre- and post-test scores. A ``High'' QLG is assigned to students who improved or remained in the high-performance group, while a ``Low'' QLG is given to those who dropped to a lower group or stayed in low/medium groups (Figure \ref{fig:qlg}). This helped identify students who benefited most from our ITS. Across the five semesters, this procedure yields 94 expert trajectories: 33 from S21, 14 from S22, 25 from S24, 13 from F24, and 9 from S25. The remaining student trajectories that were not selected as experts are treated as imperfect, sub-optimal trajectories (\textbf{I}). In addition to expert-only(\textbf{E}) training, we consider a mixed-quality setting that includes both expert and imperfect, sub-optimal student trajectories, denoted as \textbf{E+I}.  

\vspace{-0.1in}
\subsubsection{Ranking Signals:} 
HALIDE relies on the availability of a meaningful \textbf{ranking signal} to modulate the influence of heterogeneous student demonstrations during AL. Unlike the reward function in RL, which assigns an \emph{absolute} value to states or actions within a single trajectory, ranking in AL encodes \emph{comparative information across demonstrations}. In HALIDE, this comparative signal, denoted as $w_i$ (Section~\ref{sec:method}, Figure~\ref{fig:overview_ranked}), determines the relative influence of student $i$’s demonstrated pedagogical decisions during policy induction.

In this work, we define ranking using Normalized Learning Gain (NLG), a well-established measure of student learning that accounts for differences in incoming competence. NLG captures how much a student improves relative to their initial knowledge, enabling fair comparison across students with heterogeneous prior preparation. Formally, NLG is defined as $\frac{posttest-pretest}{\sqrt{1-pretest}}$, where 1 is the maximum score for both pre- and post-test. To ensure stability across cohorts and semesters, NLG values are standardized across all semesters to obtain $NLG_z$ and the  final ranking weight is then computed as
$w_{i} = \sigma(\alpha . NLG_z)$. where $\sigma(\cdot)$ denotes the sigmoid function and $\alpha$ controls the sharpness of the weighting.
This formulation maps relative learning gains to \textbf{smooth, bounded weights}, allowing HALIDE to emphasize demonstrations from students who exhibit stronger learning outcomes while retaining informative behavior from suboptimal trajectories. Importantly, this ranking signal functions as a \emph{soft structural prior} rather than a hard filter, enabling robust learning from  mixed-quality student demonstrations.


\vspace{-0.1in}
\subsubsection{Experiment Settings}
We compare \textbf{HALIDE} against a set of competitive apprenticeship learning (AL) baselines and structured ablations designed to isolate the effects of hierarchy, demonstration quality modeling, and data heterogeneity on a challenging task of predicting future semester students' actions using past ones.

\noindent\textbf{1) Flat AL Baselines:}
Behavioral Cloning (BC) \cite{gleave2022imitation} learns a policy via supervised learning on state--action pairs from expert demonstrations. 
EDM \cite{jarrett2020edm} is a state-of-the-art strictly offline apprenticeship learning method that assumes a single stationary reward function and matches expert state--action occupancy distributions.

\noindent\textbf{2) Hierarchical AL Baselines:}
We include an offline adaptation of Hierarchical Inverse Reinforcement Learning (AHIRL) \cite{krishnan2016hirl}, which reuses segment-level clustering and local reward structure but replaces online RL with EDM-based policy induction to ensure fair comparison in a fully offline setting.
\THEMES is a prior hierarchical apprenticeship learning framework that partitions trajectories into temporally coherent sub-trajectories and induces cluster-specific policies to model evolving pedagogical strategies over time. 

\noindent\textbf{3) Quality-Aware and Data-Ablations:}
Methods denoted with the subscript \textbf{W} incorporate continuous demonstration quality signals to modulate the influence of demonstrated decisions during policy induction, without altering the underlying policy or hierarchical structure.
This includes EDM\textsubscript{W} in both expert-only (E) and mixed-quality (E+I) settings. \textbf{HALIDE\(_0\)} serves as a data ablation of HALIDE, which augments hierarchical learning with quality-aware weighting while using expert demonstrations only.
Another ablation, \textbf{HALIDE\(_1\)}, jointly leverages expert and sub-optimal demonstrations (E+I), enabling the model to exploit increased state-space coverage while using uniform ranking signals.


\vspace{-0.2in}
\subsubsection{Evaluation Protocol and Metrics}
All methods are evaluated using semester-based temporal cross-validation (4-fold) which is a much stricter approach than standard cross-validation \cite{mao2020time}. Models are trained on one or more past semesters and evaluated on a future semester, and results are aggregated across multiple such folds. We report Accuracy, Recall, Precision, F1 score, Area Under the ROC Curve (AUC), Area Under the Precision--Recall Curve (APR), and Jaccard score. We emphasize F1 score, AUC, APR and Jaccard as primary metrics, as they are robust to class imbalance and better capture decision-level agreement. Reported results correspond to the mean and standard deviation across temporal folds, and statistical significance is assessed using non-parametric tests where appropriate.

\section{Results}
\label{sec:results}
\vspace{-0.1in}

Table~\ref{tab:result_xsem_ranked} reports the mean and standard deviation of semester-based temporal cross-validation results across the different methods, with the best-performing baselines and ablations highlighted in bold, and the overall best results are marked with $^{*}$. 

\renewcommand{\arraystretch}{1.3} 
\setlength{\tabcolsep}{1.2 pt}
\begin{table*}
\vspace{-0.3in}
\caption{\centering Semester-based temporal cross-validation results (mean $\pm$ std)}
\label{tab:result_xsem_ranked}
\small
\centering
\begin{tabular}{c|c|ccc||cc|cc} \hline
Data & Methods & Acc & Rec & Prec & F1 & AUC & APR & JAC \\ \hline

\multirow{5}{*}{E}
& BC \tiny{\cite{gleave2022imitation}} 
& $.434_{\pm .08}$ & $.434_{\pm .08}$ & $.490_{\pm .17}$ & $.453_{\pm .11}$ & $.488_{\pm .08}$ & $.490_{\pm .18}$ & $.314_{\pm .10}$ \\

& EDM \tiny{\cite{jarrett2020strictly}}
& $.673_{\pm .09}$ & $.673_{\pm .09}$ & $.754_{\pm .06}$ & $.695_{\pm .07}$ & $.766_{\pm .13}$ & $.755_{\pm .07}$ & $.548_{\pm .08}$ \\


& AHIRL \tiny{\cite{krishnan2016hirl}}
& $.586_{\pm .02}$ & $.586_{\pm .02}$ & $.666_{\pm .11}$ & $.602_{\pm .05}$ 
& $.716_{\pm .06}$ & $.658_{\pm .13}$ & $.443_{\pm .05}$ \\

& \THEMES
& $\mathbf{.724_{\pm .07}}$ & $\mathbf{.724_{\pm .07}}$ & $\mathbf{.784_{\pm .05}}$ 
& $\mathbf{.742_{\pm .05}}$ & $\mathbf{.803_{\pm .11}}$ & $\mathbf{.787_{\pm .06}}$ & $\mathbf{.603_{\pm .07}}$ \\ \hline

\multirow{2}{*}{E}
& EDM\textsubscript{W}
& $.674_{\pm .09}$ & $.674_{\pm .09}$ & $.755_{\pm .06}$ & $.696_{\pm .07}$ 
& $.769_{\pm .13}$ & $.757_{\pm .07}$ & $.550_{\pm .08}$ \\

& HALIDE$_0$
& $\mathbf{.738_{\pm .08}}$ & $\mathbf{.738_{\pm .08}}$ & $\mathbf{.785_{\pm .06}}$ 
& $\mathbf{.752_{\pm .06}}$ & $\mathbf{.813_{\pm .11}}$ & $\mathbf{.809^{*}_{\pm .06}}$ & $\mathbf{.617_{\pm .08}}$ \\ \hline

\multirow{2}{*}{E+I}
& EDM
& $\mathbf{.705_{\pm .08}}$ & $\mathbf{.705_{\pm .08}}$ & $\mathbf{.768_{\pm .06}}$ 
& $\mathbf{.723_{\pm .07}}$ & $\mathbf{.774_{\pm .14}}$ & $\mathbf{.767_{\pm .07}}$ & $\mathbf{.582_{\pm .09}}$ \\ 

& HALIDE$_1$
& $.698_{\pm .07}$ & $.698_{\pm .07}$ & $.766_{\pm .06}$ & $.716_{\pm .06}$ 
& $.769_{\pm .14}$ & $.756_{\pm .06}$ & $.573_{\pm .08}$ \\ \hline

\multirow{2}{*}{E+I}
& EDM\textsubscript{W}
& $.697_{\pm .09}$ & $.697_{\pm .09}$ & $.768_{\pm .06}$ & $.716_{\pm .08}$ 
& $.773_{\pm .14}$ & $.762_{\pm .07}$ & $.575_{\pm .10}$ \\

& HALIDE
& $\mathbf{.776^{*}_{\pm .07}}$ & $\mathbf{.776^{*}_{\pm .07}}$ & $\mathbf{.793^{*}_{\pm .08}}$ 
& $\mathbf{.782^{*}_{\pm .08}}$ & $\mathbf{.827^{*}_{\pm .08}}$ & $\mathbf{.801_{\pm .07}}$ & $\mathbf{.658^{*}_{\pm .10}}$ \\ \hline

\end{tabular}
\vspace{-0.2in}
\end{table*}

Across all evaluation metrics, the proposed \textbf{HALIDE} framework achieves the strongest overall performance, significantly outperforming all flat and hierarchical baselines. In particular, \textbf{HALIDE} trained on expert and sub-optimal data (E+I) attains the highest AUC (0.827) and Jaccard score (0.658), indicating superior discrimination ability and decision-level agreement under substantial cross-semester distribution shift.

Among flat apprenticeship learning baselines trained on expert data (E), \textbf{EDM} consistently outperforms \textbf{BC}, confirming that matching occupancy measures under a learned reward is more effective than direct behavioral cloning. However, flat methods remain limited by their assumption of a single stationary objective and their inability to capture temporal abstraction or evolving pedagogical strategies driven by student agency.

Comparing hierarchical approaches under expert-only data (E), \textbf{\THEMES} improves over flat baselines, demonstrating the benefit of time-aware sub-trajectory partitioning and hierarchical policy induction. However, introducing heterogeneous data without quality modeling is insufficient: \textbf{HALIDE$_1$}, trained on mixed data (E+I), does not improve over expert-only \textbf{THEMES}. This suggests that while mixed-quality data expands state-space coverage, unweighted hierarchical learning allows sub-optimal decisions to exert equal influence, diluting informative signals and limiting robustness.

\begin{figure}
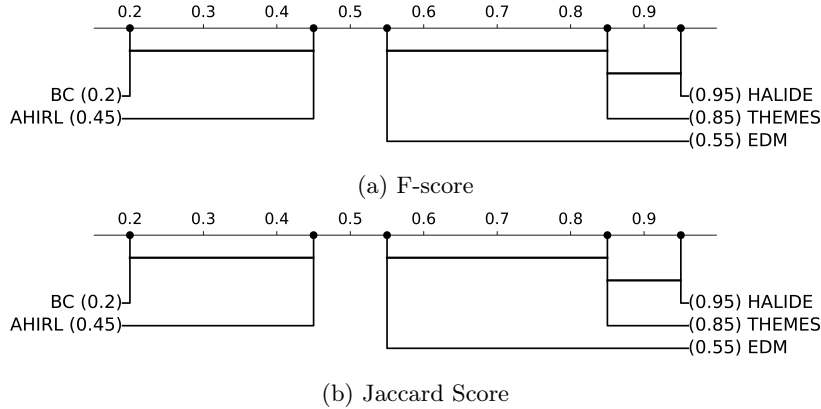

\vspace{-0.13in}
  \centering
  \begin{subfigure}{0.9\textwidth}
    \centering
    \includegraphics[width=\linewidth]{figures/cd_fscore_camera.png}
    \caption{F-score}
    \label{subfig:cd_fscore}
  \end{subfigure}
  \hfill
  \begin{subfigure}{0.9\textwidth}
    \centering
    \includegraphics[width=\linewidth]{figures/cd_jaccard_camera.png}
    \caption{Jaccard Score}
    \label{subfig:cd_jac}
  \end{subfigure}
  \caption{Critical difference diagram for F-score and Jaccard}
  \label{fig:cd_plot}
  \vspace{-0.2in}
\end{figure}

In contrast, incorporating demonstration quality via ranking signal yields consistent gains. Under expert-only data, \textbf{HALIDE$_0$} outperforms its unweighted counterpart, indicating that even demonstrations labeled as "expert" exhibit meaningful step-level variation in reliability. This challenges the standard apprenticeship learning assumption that expert trajectories are uniformly optimal and highlights the importance of modeling intra-trajectory quality differences in human-centric domains.

The largest gains emerge in the mixed-quality regime. While flat and unweighted hierarchical methods struggle to benefit from heterogeneous demonstrations, \textbf{HALIDE} effectively leverages both expert and sub-optimal data (E+I). By attenuating the influence of low-impact or erroneous decisions while preserving useful routine actions, HALIDE exploits increased state-space coverage without inheriting sub-optimal biases. This selective reuse of heterogeneous data enables HALIDE to substantially outperform both \textbf{\THEMES} and weighted flat baselines in all key metrics.

The critical difference (CD) diagrams \cite{conover1979comparison} for F1 and Jaccard scores, shown in Figure~\ref{fig:cd_plot}, summarize pairwise statistical comparisons based on the Conover– Friedman test ($p = 0.05$) with Holm correction across semester-based folds. Due to the limited number of folds ($n=4$), the test is conservative, and several methods form overlapping equivalence groups despite clear differences in mean performance. Nevertheless, HALIDE consistently attains the best average rank and is clearly separated from weaker baselines such as BC and AHIRL. The absence of separation between some intermediate methods reflects limited statistical power rather than identical performance.

Taken together, these results support the central hypothesis of this work: \emph{hierarchical apprenticeship learning benefits from explicitly ranking demonstrations by quality, particularly in realistic educational settings where behavior is heterogeneous, imperfect, and only near-optimal}. 
HALIDE reconciles robustness with data efficiency by modulating the influence of demonstrated decisions, enabling principled reuse of mixed-quality student data while preserving the temporal and strategic structure induced by hierarchical modeling with evolving rewards.
\section{Discussion, Caveats,  and Conclusions}
\label{sec:discussion}
\vspace{-0.1in}

This work addresses a key limitation of AL in student-centric educational settings: student pedagogical demonstrations are inherently heterogeneous in quality, even among high-performing learners. Students differ in prior knowledge, self-regulation, and metacognitive skills, and their decision quality fluctuates over time, violating the standard AL assumption that expert demonstrations are uniformly reliable. Our results show that treating all demonstrations equally injects noise into policy induction, limiting robustness—especially under cross-semester distribution shifts and mixed-quality data.

HALIDE operationalizes a principle long established in mixed-initiative ITS research—student agency must be balanced with selective modulation of influence. Rather than discarding imperfect trajectories or uniformly weighting demonstrations, HALIDE integrates continuous demonstration quality signals within a hierarchical apprenticeship learning framework. This allows useful routine decisions from sub-optimal trajectories to contribute while attenuating the influence of low-impact or erroneous actions. Empirically, this integration is critical: unweighted hierarchical learning fails to benefit from additional heterogeneous data, whereas HALIDE consistently improves performance, with the largest gains observed when leveraging both expert and sub-optimal demonstrations (E+I). These results highlight the complementary roles of ranking and hierarchy, where quality-aware weighting stabilizes low-level policy induction and hierarchical structure propagates these improvements to higher-level pedagogical strategies.


Despite these contributions, this work has several limitations. First, demonstration quality is estimated at the trajectory level and propagated across steps, which provides scalability but only an approximate proxy for step-level pedagogical effectiveness. Second, all evaluations are conducted offline and within a single ITS domain, leaving open questions regarding performance under closed-loop deployment and generalization across domains or curricula. Third, the quality ranking mechanism may reflect biases present in historical student data—such as differences in prior preparation, engagement, or learning norms—which could influence policy induction if not carefully monitored.
More broadly, HALIDE relies on the availability of a relative quality ranking to modulate the influence of heterogeneous demonstrations. Although ranking is substantially weaker and easier to obtain than specifying a full reward function, its quality can affect downstream learning. Noisy or misaligned rankings—such as those correlated with short-term performance rather than pedagogical effectiveness—may attenuate the benefits of quality-aware weighting. This limitation highlights the importance of careful ranking design and validation when applying HALIDE in practice. 
Despite these caveats, our findings demonstrate that modeling demonstration quality as a graded structural signal—rather than a binary selection criterion—is essential for robust apprenticeship learning under student agency. By combining quality-aware weighting with hierarchical modeling and evolving rewards, HALIDE offers a principled path toward deploying AL in realistic, heterogeneous educational environments.

\section{Acknowledgments}
This research was supported by the NSF Grants: CAREER: Improving Adaptive Decision Making in Interactive Learning Environments(1651909), and Generalizing Data-Driven Technologies to Improve Individualized STEM Instruction by Intelligent Tutors (2013502), and Empowering Learning-by-Teaching through Reinforcement Learning and Explainable AI (2507143).
%
%
%
\bibliographystyle{splncs04}
\bibliography{reference}

\end{document}